\pdfoutput=1

\documentclass[11pt]{article}

\usepackage{acl2024}

\usepackage{times}
\usepackage{multirow}
\usepackage{latexsym}
\usepackage[inline,shortlabels]{enumitem}
\usepackage[T1]{fontenc}
\usepackage{float}
\restylefloat{table}

\usepackage[utf8]{inputenc}

\usepackage{microtype}

\usepackage{inconsolata}

\usepackage{booktabs}
\usepackage{adjustbox}
\usepackage{graphicx}
\usepackage{fdsymbol}
\usepackage{amsmath}
\usepackage{xcolor}
\usepackage{pifont}
\usepackage{color,soul}

\newcommand{\Figref}[1]{Figure~\ref{#1}}  

\newcommand{\Tabref}[1]{Table~\ref{#1}}
\newcommand{\secref}[1]{Section~\ref{#1}}

\newcommand{\appref}[1]{Appendix~\ref{#1}}

\newcommand{\cmark}{\ding{51}}%
\newcommand{\xmark}{\ding{55}}%

\newcommand{\Hl}[2][\empty]{%
\ifx#1\empty
\else
\sethlcolor{#1}%
\fi
\hl{#2}}

\newcommand\mname{MAC}

\title{Language Models Guidance with Multi-Aspect-Cueing: \\ A Case Study for Competitor Analysis}

\author{Amir Hadifar, Christopher Ochs, Arjan Van Ewijk \\
  Nokia Bell Labs \\
  \texttt{firstname.lastname@nokia-bell-labs.com} \\
 }

\begin{document}
\maketitle
\begin{abstract}
Competitor analysis is essential in modern business due to the influence of industry rivals on strategic planning. It involves assessing multiple aspects and balancing trade-offs to make informed decisions.
Recent Large Language Models (LLMs) have demonstrated impressive capabilities to reason about such trade-offs but grapple with inherent limitations such as a lack of knowledge about contemporary or future realities and an incomplete understanding of a market's competitive landscape.
In this paper, we address this gap by incorporating business aspects into LLMs to enhance their understanding of a competitive market.
Through quantitative and qualitative experiments, we illustrate how integrating such aspects consistently improves model performance, thereby enhancing analytical efficacy in competitor analysis.

\end{abstract}

\section{Introduction}
Businesses closely monitor their rivals through \textit{competitor analysis} to anticipate moves and trends.
They systematically analyze various sources, such as news and social media, to uncover insights into a competitor's strategic plans, positioning strategies, and potential shifts in competitive advantage within an industry \cite{adom2016competitor}. 
However, performing such a manual is laborious due to the rapid rate of content creation.




Natural Language Processing (NLP) techniques can provide insights and optimize analysis processes in critical ways such as: \begin{enumerate*}[(i)]
\item find potential rivals \cite{cao2023companykg}
\item categorize and prioritize media sources \cite{arslan2023leveraging}
\item aid in generating reports or summaries of findings \cite{ma2023insightpilot}.
\end{enumerate*}
Nevertheless, businesses are still confronted with challenges such as identifying relevant signals aligned with their unique needs, integrating prior knowledge into NLP systems, and finding optimal balances among multiple criteria.

Recent Large Language Models (LLMs) have shown remarkable problem-solving capabilities \cite{bubeck2023sparks}. Through conditioning on specific user instructions, known as \textit{prompts}, these models can be customized to effectively address complex reasoning tasks \cite{wei2022chain,kojima2022large}.
Motivated by this, we propose a technique to decompose a competitive landscape into a few key \textit{aspects}, thereby steering LLM toward business preferences in a specific market.
These aspects can be any arbitrary features appended to the original prompt to more accurately articulate a specific user intent. For instance, within the news excerpt: ``\textbf{Telecom giant}, announces strategic partnership to accelerate the \textbf{5G Core}'' the bold segments exemplify such aspects.

Through experiments, we demonstrate the usefulness of integrating multiple aspects, sourced from an enterprise's pre-existing knowledge, into a prompt.
Using examples from the telecommunications industry, we illustrate the effectiveness of Multi-Aspect Cueing (\mname) and showcase the distinct contributions of individual aspects to the model predictions, offering insights into how each aspect influences the model's decision-making process (\secref{sec:experiment_1}). Additionally, we show that integrating aspects significantly enhances the memorization ability of LLM (\secref{sec:experiment_2}), thereby addressing the challenge of discrepancies between models trained on past data and the need to analyze dynamic competitive landscapes. Finally, we provide a quantitative comparison of {\mname} against several baselines using data from several different domains (\secref{sec:experiment_3}).

\section{Related work}
\label{sec:related_work}
Significant research has been devoted to leveraging NLP for competitor analysis. 
\citet{jin2016minimizing} used patents and litigation cases to forecast potential legal harm from rivals. \citet{menon2018you} analyzed 10-K reports to study business changes and competitive positioning. \citet{zhang2020large} examined job transitions among companies using online data. 
\citet{lee2021extraction} identified competitiveness factors from online product reviews. \citet{cao2023companykg} introduced a knowledge graph for competitor retrieval and company similarity analysis.


LLMs have showcased notable capabilities in competitor analysis and associated tasks.
These models can be tailored to a downstream task through various methods, including finetuning \cite{liu2019roberta}, and prompting, \cite{brown2020language}. 
\citet{wu2023bloomberggpt,yang2023fingpt} finetuned LLMs for the financial domain and benchmarked their capabilities on sentiment analysis, classification, and named entity recognition tasks. 
\citet{ma2023insightpilot,weng2024insightlens} utilized prompting to streamline insight extraction by transforming natural questions posed by users into a series of analytical actions and then generating insights.
These prompting techniques were further improved with additional refinement. For instance, Chain-of-Thought (CoT) approaches \cite{wei2022chain,kojima2022large} break down complex reasoning tasks into smaller subtasks to facilitate easier comprehension for LLMs.  

Although the performance gains achieved by LLMs are well-established, their effectiveness in multi-aspect decision-making remains unexplored.
Some studies have investigated aspect-based scenarios \cite{kong2022multi,zhang2023recipe,sun2023text,teixeira2023enhancing,thomas2023large} but they do not consider the effect of incorporating multiple aspects, as described in this paper.

\section{Methodology}
\label{sec:methodology}
\begin{figure}[ht!]
  \centering
  \includegraphics[width=\linewidth]{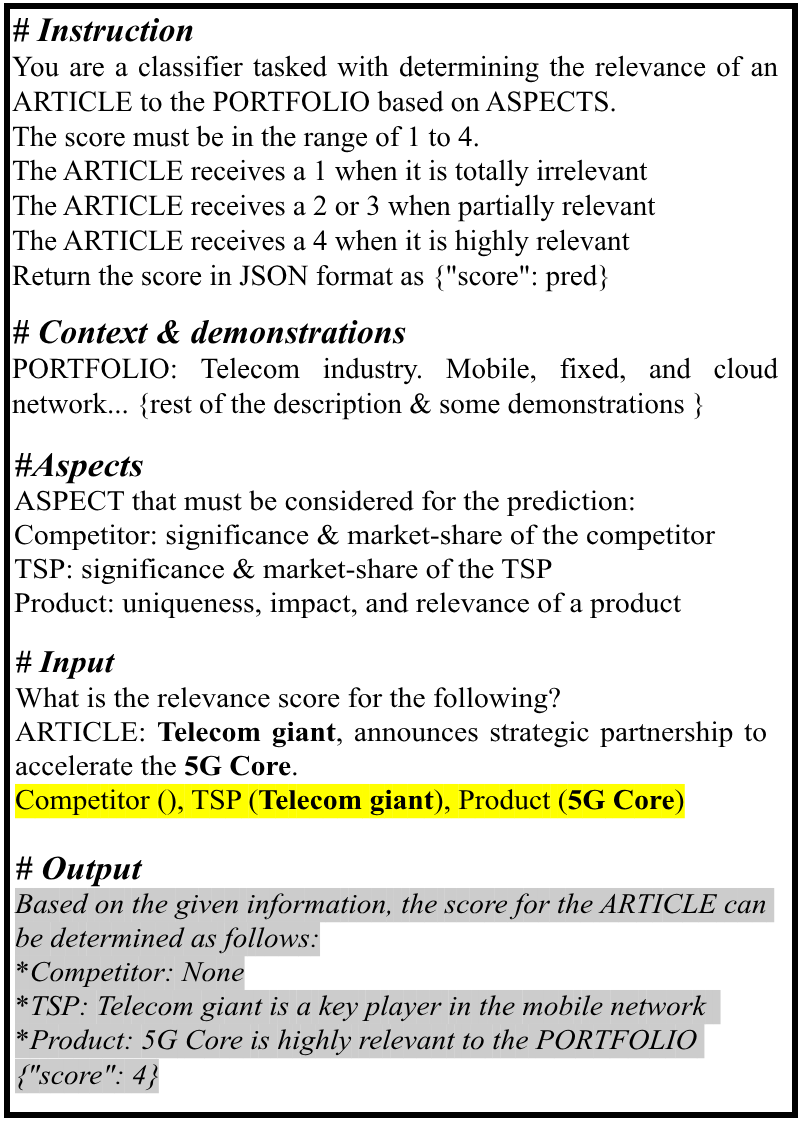}
   \caption{Multi-Aspect Cueing (\mname) example for competitor analysis. \Hl[yellow!85]{Highlighted} words are aspects, and \Hl[gray!40]{shaded} are generated text by Llama2.}
  \label{fig_prompt}
\end{figure}

We propose Multi-Aspect Cueing (\mname), a framework for directing LLMs by integrating multiple aspects that guide the LLMs to generate desired responses.
\Figref{fig_prompt} provides an overall schema of {\mname}, illustrating its use in obtaining relevance scores for an article related to the telecommunications market. 
Utilizing {\mname} for this use case can identify company-specific and business-critical information (e.g., partnerships and product developments).
In the following paragraphs, we will briefly explain the different components of {\mname}.




\textit{Instruction:} task instruction, offering general guidance for the LLM. The instruction can also determine the output format, range, or any other contextual information that is relevant to the task. 

\textit{Context \& demonstrations:} relevant information about the topic or task. This information helps the LLMs better understand the purpose and significance of the task it is being asked to perform. Context can also be accompanied by a few demonstrations or exemplars. 

\textit{Aspects:} elements that the LLM should focus on are determined by explicitly considering the user's intent and decomposing it into multiple aspects. These aspects not only facilitate the division of complex analyses into simpler components but also incorporate domain-specific information that might be underrepresented or misrepresented in the training data of LLMs. 

\textit{Input:} the item (e.g., article) that has to be annotated by the LLM. Along with the article, corresponding values for each of the aspects that we found in the article are included. These values can be sourced e.g., from a pre-existing knowledge base, pattern, or heuristics.
Through the incorporation of these values, we guide the LLM toward generating desired outcomes.

\textit{Output:} LLM's output, reflecting its understanding and processing of the provided information. 


\section{Experimental Results}
In this section, we will present our experimental results. 
We start by visualizing the contribution of aspects on the LLM prediction (\secref{sec:experiment_1}). Additionally, we scrutinize the memorization ability of LLM with and without incorporating aspects for a telecommunication use case (\secref{sec:experiment_2}). These two experiments serve as a sanity check that LLM can effectively utilize the input aspects during its reasoning process. 
Finally, we evaluated {\mname} on three datasets covering disjoint topic domains: telecommunications, recipes, and finance (\secref{sec:experiment_3}). {\mname} was applied on Llama2 \cite{touvron2023llama} for all experiments (see details in \appref{app:llama2_hyperparameters}).

\subsection{Aspects Attribution}
\label{sec:experiment_1}
In this experiment, we visually demonstrate the contribution of the input aspects to the LLM prediction.
This aids in understanding the model's decision-making process and the influence of different aspects on output sequences.
To compute the contribution of these aspects, we apply Shapley Value Sampling \cite{castro2009polynomial},\footnote{We used existing implementation at: \url{https://github.com/pytorch/captum}} which involves permuting the aspects, sequentially adding them to a baseline (i.e., without aspects), and measuring the output change with each addition. 
Utilizing the {\mname} template (\secref{sec:methodology}), tailored with 10 exemplars, we generate the most probable subsequent tokens to classify an input article according to its relevance to the telecommunication industry. The classification scale ranges from 1 to 4, where 1 denotes irrelevance and 4 indicates high relevance. 

\begin{figure}
  \centering
  \includegraphics[width=\linewidth]{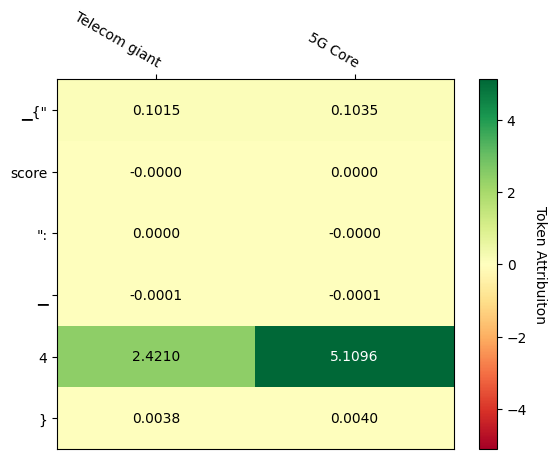}
   \caption{A visual example of aspect attributions using Shapely Value Sampling \cite{castro2009polynomial}. Both aspects positively contribute to the predicted score.}
  \label{fig_score}
\end{figure}

\Figref{fig_score} shows the token level attributions on specified output sequence (i.e., \{"score":4\}) for news excerpt: ``\textbf{Telecom giant}, announces strategic partnership to accelerate the \textbf{5G Core}'', with respect to target aspects (bold segments). As shown, the aspects significantly contributed to the predicted score. For example, ``5G Core'' had contributions of $5.1096$, likely due to more proximity to the telecommunication compared to ``Telecom giant''. This example suggests that the input aspects in {\mname} contribute to output tokens (see more examples in  \appref{app:aspect_contribution}).

\begin{table*}[tb]
    \centering \small
    \renewcommand{\arraystretch}{1.2}
    \footnotesize
    \begin{tabular}{l cc cc cc}
    \toprule
      
    & \multicolumn{2}{c}{\textbf{InHouse}} 
    & \multicolumn{2}{c}{\textbf{Recipe-MPR}} 
    & \multicolumn{2}{c}{\textbf{FiQA SA}}
    \\

    \cmidrule(l{3pt}r{3pt}){2-3}
    \cmidrule(l{3pt}r{3pt}){4-5}
    \cmidrule(l{3pt}){6-7}
    
    \multicolumn{1}{c}{\textbf{Model}} &  
    
    \multicolumn{1}{c}{\textbf{ACC}} & 
    \multicolumn{1}{c}{\textbf{F1$_{macro}$}} & 
    
    \multicolumn{1}{c}{\textbf{ACC}} & 
    \multicolumn{1}{c}{\textbf{F1$_{weighted}$}} & 
    
    \multicolumn{1}{c}{\textbf{ACC}} & 
    \multicolumn{1}{c}{\textbf{F1$_{weighted}$}}\\

    \midrule
    0Shot-TC \cite{yin2019benchmarking} & 7.3$\pm$0.8 & 1.6$\pm$0.1 & 44.7$\pm$3.4 & 28.1$\pm$3.6 & 69.7$\pm$0.5 & 64.6$\pm$0.7 \\
    
    FT-Roberta \cite{liu2019roberta} & 80.6$\pm$0.4 & 54.1$\pm$0.5  & 47.3$\pm$2.1 & 47.2$\pm$2.3 & 83.3$\pm$1.4 & 83.1$\pm$0.7 \\
    
    LLama2-70B \citep[5-shot;][]{touvron2023llama} & 36.3$\pm$5.1 & 29.4$\pm$4.2 & 44.8$\pm$2.1 & 44.1$\pm$2.2 & 78.1$\pm$1.5 &               80.5$\pm$1.5 \\ 
    
             \ + CoT \cite{kojima2022large}  & 67.8$\pm$1.9 & 47.9$\pm$2.1 & 47.8$\pm$2.8 & 46.9$\pm$2.1 & 80.3$\pm$1.3 & 80.5$\pm$1.4\\
             
             \ + \textbf{\mname}  & 71.2$\pm$0.6 & 46.8$\pm$2.2 & 53.2$\pm$2.3 & 50.2$\pm$1.9 & 85.3$\pm$1.3 & 85.9$\pm$1.3 \\

    Llama2-70B \citep[10-shot;][]{touvron2023llama} & 40.8$\pm$5.3 & 30.3$\pm$4.7 & 48.4$\pm$3.7 & 45.4$\pm$4.0 & 81.0$\pm$1.4 & 82.6$\pm$1.2 \\ 
    
             \ + CoT \cite{kojima2022large}  & 78.7$\pm$2.3 & 52.3$\pm$0.9 & 50.0$\pm$2.6 & 48.3$\pm$2.8 & 83.2$\pm$1.7 & 83.1$\pm$2.1\\
             
             \ + \textbf{\mname}  
             & 82.2$\pm$1.4 & 56.8$\pm$1.7 & 55.1$\pm$3.0 & 54.1$\pm$2.9 & 86.2$\pm$2.0 & 87.3$\pm$2.1 \\

    
    
             

    
             
             
    \bottomrule

    \end{tabular}
    \caption{\label{tb:results} Multi-Aspect Cueing evaluation on three baseline datasets.}
\end{table*}

\subsection{Memorization with Aspect-Cueing}
\label{sec:experiment_2}
To further investigate the functionality of integrating aspects into the prompt, we evaluate the memorization ability of LLM with and without {\mname}. Here, memorization ability refers to retaining and recalling input aspects throughout the reasoning process, ensuring their presence in the output without omission (see example in \appref{app:memorization_example}). 
Memorization is important, particularly for aspects misrepresented or excluded in LLM's training data.

To this aim, we manually compiled a collection of $155$ Telecommunication Service Providers\footnote{\url{https://en.wikipedia.org/wiki/List_of_telephone_operating_companies}} (TSPs), crucial entities for connectivity in the telecom industry. Each TSP is accompanied by five news headlines where its name is precisely mentioned in the text. We categorized TSPs into four distinct categories—\textit{Highly-Frequent}, \textit{Frequent}, \textit{Less-Frequent}, and \textit{Rare} based on their frequency of appearance in news sources over a recent period (see \secref{app:memorization_frequency}). The frequency serves as a proxy for estimating the presence of TSPs in the training data of LLMs, although accurately measuring this is challenging.
Following this, Llama2 was tasked with detecting aspects, specifically the TSP names, within the provided headline. We employed {\mname} template, and tailored it with $10$ exemplars.

\begin{figure}
  \centering
  \includegraphics[width=\linewidth]{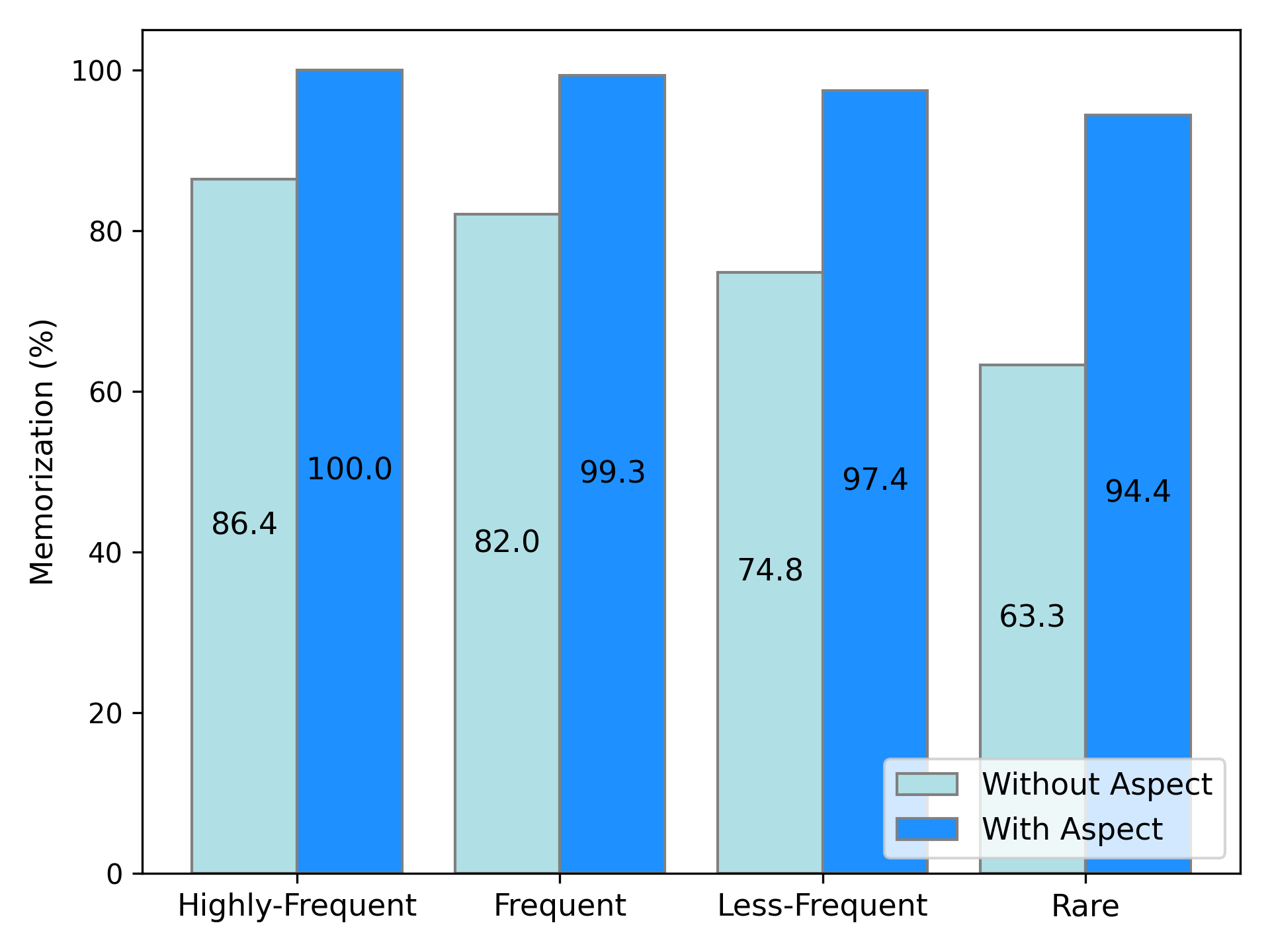}
   \caption{LLM memorization for recognizing Telecommunication Service Provider (TSP) in headlines for the varying frequency levels with/without aspect-cueing. }
  \label{fig_ood}
\end{figure}

\Figref{fig_ood} shows the memorization ability of Llama2. As shown, incorporating aspects into the prompt proves advantageous, particularly for domain-specific TSPs that may be underrepresented or misrepresented in the training data of LLama2.
Upon integrating the TSP into the prompt, a notable improvement in memorization scores is observed, with rates reaching $100\%$ and $94.4\%$ for \textit{Highly-Frequent} and \textit{Rare} TSPs, from $86.4\%$ and $63.3\%$, respectively. 
This observation suggests that aspects receiving less visibility within the pre-training datasets of LLMs, or those that exhibit contextual ambiguity, tend to benefit more from our approach. Furthermore, a lingering incongruity persists in the memorization rates pertaining to \textit{Less-Frequent} and \textit{Rare} TSP, a phenomenon that could be ascribed to knowledge conflict \cite{xu2024knowledge}.

\subsection{Multi-Aspect Cueing}
\label{sec:experiment_3}
In this section, we consider evaluating our hypothesis on two publicly available datasets Recipe-MPR \cite{zhang2023recipe}, FiQA SA \cite{maia201818}, and one proprietary dataset (denoted by InHouse) as we are not able to find a multi-aspect competitor analysis dataset suitable for our experiments (see more details about datasets in \secref{app:datasets}).
For baselines we evaluate it against zero-shot text classification \citep[0Shot-TC;][]{yin2019benchmarking}, fine-tuned roberta \citep[FT-Roberta;][]{liu2019roberta},  Llama2-70B \cite{touvron2023llama}, and Llama2-70B with CoT \cite{kojima2022large}. For Llama2 we employed $5$ and $10$ examples as a few-shot setup (see \secref{app:experimental_setup}). 

\Tabref{tb:results} summarizes the results. The integration of multiple aspects into decision-making processes enhances performance, particularly when tasks necessitate temporal knowledge or the inclusion of external information. This improvement is notably pronounced in the InHouse dataset, where the incorporation of aspects leads to substantial gains over vanilla prompting (more than $30\%$ for $5$-shot and $40\%$ for $10$-shot). 
0Shot-TC performs well on Recipe-MPR and FiQA SA, however, it struggles with the InHouse dataset that involves multifaceted criteria for prediction and output labels are ill-defined. FT-Roberta outperformed various prompting techniques, which can be attributed to the larger number of training samples it encountered during the fine-tuning process. 
The performance of Llama is further enhanced with the incorporation of CoT, MAC, and additional demonstrations. CoT and MAC improve the model's reasoning capabilities, while the provision of more examples aids the model learn to trade-off between multiple criteria effectively.

\section{Conclusion}
\label{sec:conclusion}
We investigated integrating multiple aspects using the {\mname} framework to further guide language models for competitor analysis tasks. We confirmed prior knowledge that providing guidance to the model enhances LLM's reasoning capability. Injecting aspects enhances both the memorization ability and reasoning capability of the model. We derive the aforementioned findings from different experiments.

\clearpage
\section*{Limitations}
While this study offers a valuable exploration of LLM-based competitor analysis, it
should be considered a stepping stone for more comprehensive research in this domain. Specifically:

\textit{Dataset}.
The scarcity and diversity of multi-aspect datasets pose challenges. Exploring alternative datasets, such as patent databases or financial filings like 10-K reports, could enrich our understanding of multi-aspect problems.

\textit{Version-Specific Findings}. The findings are context-specific, based on Llama2 and future updates to the model may impact results.



\bibliography{acl2024}

\appendix

\section{Llama2 Hyperparameters}
\label{app:llama2_hyperparameters}
We are executing the 16-bit quantized  \citep[GPTQ;][]{frantar2022gptq} variant of Llama2-70B on 8$\times$A5000 GPUs. The specific hyperparameters employed are detailed in \Tabref{tb:llama_hyperparameters}.

\begin{table}[ht!]
    \centering
    \renewcommand{\arraystretch}{1.2}
    \footnotesize
    \begin{tabular}{l c}
    \toprule
    
    \textbf{Hyperparameter} &

    \textbf{Value}  
    
    \\
    
    \midrule
    
       temperature    & $0.3$   \\
       top\_p & $0.95$ \\
       repetition\_penalty & $1.2$ \\
       top\_k & $50$ \\
       max\_new\_tokens & $400$ \\

    \bottomrule

    \end{tabular}
        
    \caption{Llama2-70B hyperparamters used in our experiments.}
    \label{tb:llama_hyperparameters}
\end{table}

\section{Aspect Attribution Example}
\label{app:aspect_contribution}
\begin{figure}
  \centering
  \includegraphics[width=\linewidth]{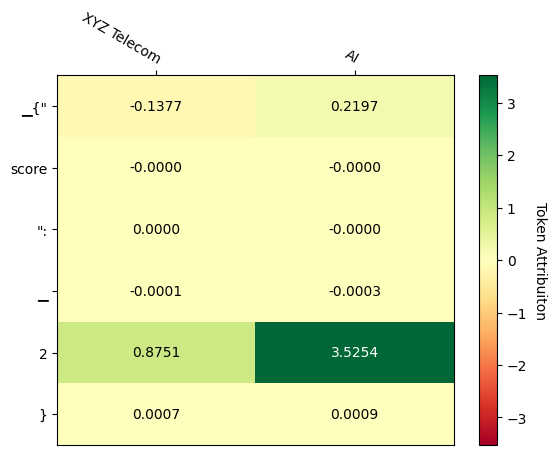}
   \caption{A visual example of aspect contributions on LLM output for news excerpt: ``\textbf{XYZ Telecom} to launch \textbf{AI} assistants for corporate clients'', with specifying target attributes.}
  \label{fig_score_xyz}
\end{figure}

\begin{figure}
  \centering
  \includegraphics[width=\linewidth]{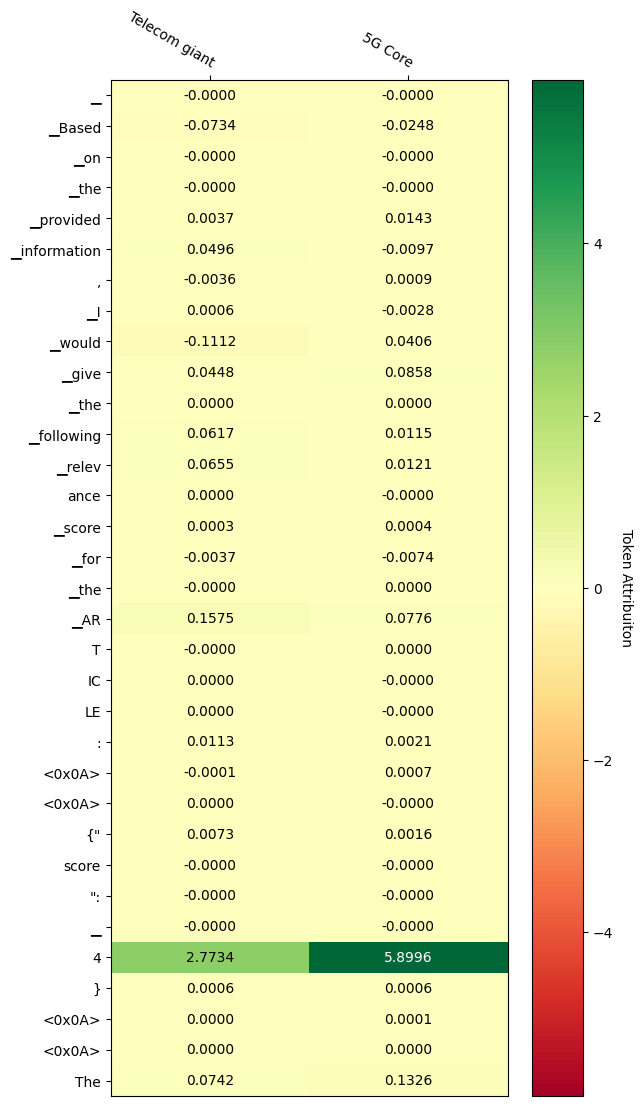}
   \caption{A visual example of aspect attributions for news excerpt: ``\textbf{Telecom giant}, announces strategic partnership to accelerate the \textbf{5G Core}'', without specifying target aspects.
   }
  \label{fig_score_full}
\end{figure}

\begin{figure}
  \centering
  \includegraphics[width=\linewidth]{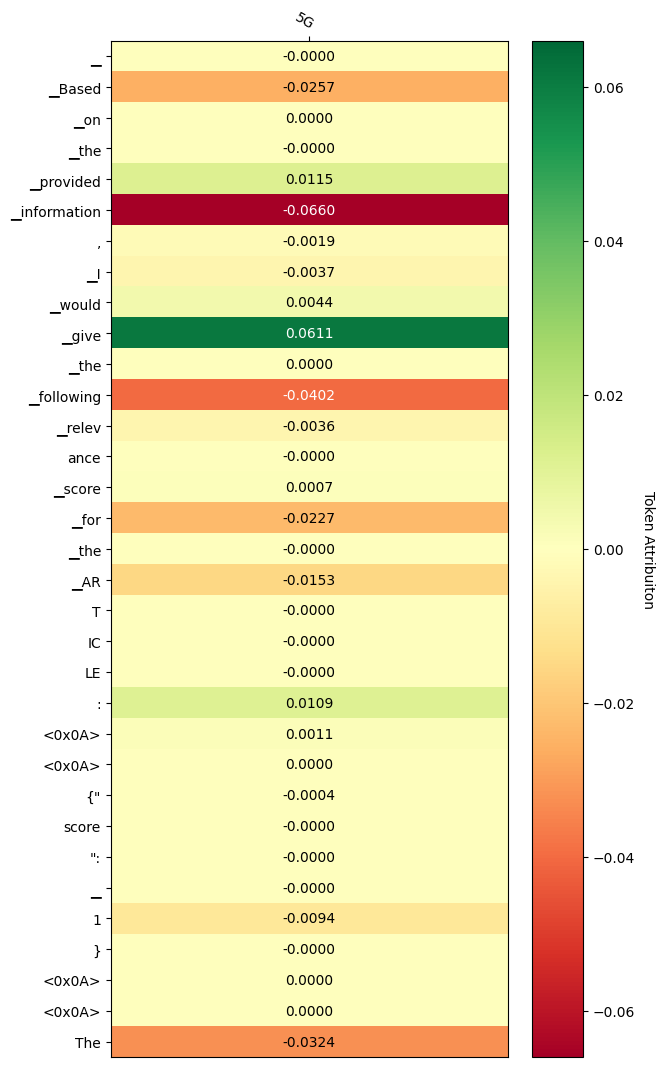}
   \caption{A visual example of aspect attributions for news excerpt: ``Mighty Squirrel expands its \textbf{5G} protein beer brand in the US'', without specifying target aspects.
   }
  \label{fig_score_full_2}
\end{figure}

In this section, we provide examples of aspect attribution for different scenarios:

\begin{itemize}
    \item \Figref{fig_score_xyz} shows the token attributions in news excerpt: ``\textbf{XYZ Telecom} to launch \textbf{AI} assistants for corporate clients'', with specifying attributions for the target outputs (i.e.,\{"score":2\}). Although, \textbf{XYZ Telecom} is not an actual Telecommunication Service Provider (TSP), the model uses it as logical and meaningful contextual information.
    
    \item \Figref{fig_score_full} shows the token attributions in the news excerpt: ``\textbf{Telecom giant}, announces strategic partnership to accelerate the \textbf{5G Core}''. No target aspect is provided in this case, so attributions are provided for the most likely decoded sequence.\footnote{The maximum number of tokens to generate is set to 33.}  

    \item \Figref{fig_score_full_2} shows the token attributions of wrongly detected aspect in the news excerpt: ``Mighty Squirrel expands its \textbf{5G} protein beer brand in the US'', where model favors its internal knowledge since the external evidence (i.e., 5G), does not provide logical and semantically meaningful information.
    
\end{itemize}

\section{LLM Memorization Example}
\label{app:memorization_example}
The memorization ability of LLM is measured with the number of times that LLM can correctly memorize the mentioned TSP from input and use it for its reasoning process. \Tabref{tb:memorization_example} shows an example of the same input with and without aspect-cueing. The input aspect (i.e., Mascom), provides a signal, in the absence of contextual knowledge, that indicates the relevance of Mascom to the telecom industry. 

\begin{table*}[ht!]
    \caption{Comparing the memorization capabilities of Llama2 with and without MAC. With aspects, the model successfully identifies `Mascom' as TSP and utilizes it in its reasoning process, whereas, without aspects, the recognition of TSP is lacking.}
    \label{tb:memorization_example}
    \centering
    \begin{tabular}{p{0.4 \linewidth} p{0.4\linewidth} 
    c
    }
\toprule

\multicolumn{1}{c}{\textbf{Input}} & 
\multicolumn{1}{c}{\textbf{Output}} &
\multicolumn{1}{c}{\textbf{Memorize}}
\\

\midrule
    



    What is the relevance SCORE for the following ARTICLE?
    
    ARTICLE: Mascom announces new price increases in 2024
     
      &\Hl[gray!40]{\textit{The article mentions Mascom, a company that is increasing its prices in 2024. However, the article does not mention any relevant competitor, TSP, or product that is relevant to the portfolio. Therefore, the score for this article would be 1, as it is not relevant to the portfolio.}

\textit{Here's the breakdown:}

\textit{* Competitor: None}

\textit{* TSP: None}

\textit{* Product: None}

\textit{\{``score'': 1\}}
}
& \xmark
\\
\midrule




What is the relevance SCORE for the following ARTICLE?
    
    ARTICLE: Mascom announces new price increases in 2024
    
    \Hl[yellow!80]{ASPECTS: Competitor (), TSP (Mascom), Product ()}
& 
\Hl[gray!40]{\textit{Based on the given information, the score for the ARTICLE can be determined as follows:}

\textit{* Competitor: None}

\textit{* TSP: Mascom is a TSP}

\textit{* Product: None}

\textit{\{``score'': 2\}}
}
&
\cmark

\\
     
    \bottomrule

    \end{tabular}
\end{table*}

\section{TSP Categorization}
\label{app:memorization_frequency}
We categorized Telecommunication Service Providers (TSPs) into four distinct categories based on the frequency of their appearances in news headlines over the past five years.

The categorization is as follows:

\begin{enumerate}
    \item \textit{Rare}: Providers appearing fewer than 10 times.  
    
    \item \textit{Less-Frequent}: Providers appearing between 10 and 1,000 times.
    
    \item \textit{Frequent}: Providers appearing between 1,000 and 10,000 times.
    
    \item \textit{Highly-Frequent}: Providers appearing more than 10,000 times.
\end{enumerate}

\section{Baseline Datasets}
\label{app:datasets}
This section provides an overview of the datasets that were used in experiments (\secref{sec:experiment_3}. These datasets include:

\begin{itemize}
\item \textbf{FiQA SA} is a multi-aspect sentiment analysis dataset for financial news. Analyzing sentiments offers insights into the perceptions of the market towards competitor products or services.
Since FiQA SA is annotated on a continuous scale, we discretize the data into a classification setup as \cite{wu2023bloomberggpt}.

\item \textbf{Recipe-MPR} is a multi-aspect multiple-choice question-answering dataset for food recommendations. This type of analysis can give restaurants a competitive advantage over their rivals by enhancing the user experience by capturing their food preferences. We conduct our experiment in a monolithic setting as it did in \cite{zhang2023recipe}.

\item \textbf{InHouse} is our in-house competitor-analysis collection which contains 1993 news items related to the telecom industry that was gathered between January and April 2023. We filtered news using keywords such as 6G, 5G, 4G, wireless, mobile network, telecommunication, etc, limited to English news.
The news is manually annotated by human experts according to their importance, impact, and relevance on the telecom industry in the range of 1-4, where 1 denotes irrelevance and 4 denotes high-relevance. The distribution of each category can be seen in \Tabref{tb:inhouse_stat}. Data annotation, as well as curation, was performed by the authors of the paper. Data annotation involved two authors independently assessing the articles, with a third author resolving any discrepancies to determine the definitive categorization. The inter-rater agreement was assessed using Fleiss' Kappa score \cite{fleiss1971measuring}, resulting in a value of $0.43$.

\end{itemize}

\begin{table}[ht!]
    \centering
    \renewcommand{\arraystretch}{1.2}
    \footnotesize
    \begin{tabular}{l c }
    \toprule
    
    \textbf{Category} &

    \textbf{Size}  
    
    \\
    
    \midrule
    
       high-relevance    & 50   \\
       moderate-relevance & 137 \\
       less-relevance & 370 \\
       irrelevance & 1436 \\

    \bottomrule

    \end{tabular}
        
    \caption{Statistics of our InHouse dataset.}
    \label{tb:inhouse_stat}
\end{table}

It is worth noting, that the first two datasets are among the few available ones that offer multi-aspect analysis.




\section{Experimental Setup}
\label{app:experimental_setup}
 Due to the lack of predefined test-train splits in baseline datasets, we hold out 10\% of the data for testing, while the remaining part is used for training. Reported results represent averages from five random seeds. Evaluation metrics were customized for each dataset: Accuracy and Weighted F1-score, typical for FIQA SA and MPR-recipe datasets, were used. Given the class imbalance in the InHouse dataset, Accuracy and F1-macro were selected to ensure a balanced assessment across all classes.

\Tabref{tb:finetune_hyperparams} also show the hyperparameters that are used for fine-tuning the Roberta-Base \cite{liu2019roberta}.
\begin{table}[h!]
    \centering
    \renewcommand{\arraystretch}{1.2}
    \footnotesize
    \begin{adjustbox}{max width=\textwidth}
    \begin{tabular}{ll c c }
    \toprule
     & 
     \multicolumn{3}{c}{\textbf{Datasets}} \\
    \cmidrule(l{3pt}r{3pt}){2-4}
    \textbf{Hyperprameter} &
    \multicolumn{1}{c}{\textbf{InHouse}}  &  
    \multicolumn{1}{c}{\textbf{Recipe-MPR}} &
    \multicolumn{1}{c}{\textbf{FIQA SA}}\\

    \midrule
    
       learning-rate     &  $5\mathrm{e}{-5}$  &  $1\mathrm{e}{-5}$  & $4\mathrm{e}{-5}$  \\
       batch-size &      $16$  &    $8$  &  $16$    \\
       training-epoch &      $10$ &     $20$ & $10$ \\
       weight-decay & $0.01$  &    $0.0$ & $0.01$ \\
    
    \bottomrule

    \end{tabular}
    \end{adjustbox}
        
    \caption{The hyperparameters used for fine-tuning roberta-base on three baseline datasets.}
    \label{tb:finetune_hyperparams}
\end{table}
We employed the same set of demonstrations for all prompt-based approaches, and the reasonings for CoT are crafted manually for each dataset.

\end{document}